\documentclass[a4paper]{article}

\usepackage{INTERSPEECH2018}
\usepackage{bm}
\usepackage{comment}	
\usepackage{amssymb}
\usepackage{bigints}  
\usepackage{float}
\usepackage{adjustbox}
\usepackage{amsmath}

\DeclareMathOperator*{\argmin}{arg\,min}

\title{Combining Natural Gradient with Hessian Free Methods\\ for Sequence Training}
\name{Adnan Haider, Philip C. Woodland   \thanks{Adnan Haider has been funded by the IDB Cambridge International Scholarship.  
Many thanks to Dr Chao Zhang for helping us build the  SGD baseline models used for this work.}}

\address{
  Cambridge University Engineering Dept., Trumpington St., Cambridge, CB2 1PZ U.K.}
\email{{\{mah90, pcw\}@eng.cam.ac.uk}}

\begin{document}

\maketitle
\begin{abstract}
This paper presents a new optimisation approach to train Deep Neural Networks (DNNs) with discriminative sequence criteria. At each iteration, the method combines information from the Natural Gradient (NG) direction with local curvature information of the error surface that enables better paths on the parameter manifold to be traversed. The method has been applied within a Hessian Free (HF) style optimisation framework to sequence train both standard fully-connected DNNs and Time Delay Neural Networks as speech recognition acoustic models. The efficacy of the method is shown using experiments on a Multi-Genre Broadcast (MGB) transcription task and neural networks using sigmoid and ReLU activation functions have been investigated. It is shown that  for the same number of updates this proposed approach achieves larger reductions in the word error rate (WER) than both NG and HF, and also leads to 
a lower WER than  standard stochastic gradient descent.




\end{abstract}


\section{Introduction}
In recent years, Deep Neural Networks (DNNs) embedded within a hybrid Hidden Markov model (HMM) framework have become the standard approach to Automatic Speech Recognition (ASR) tasks \cite{Hinton2012}. 
While the  structure of such DNN models gives rich modelling capacity and yields good performance, it also creates complex dependencies between the parameters which can make learning difficult via first order Stochastic Gradient Descent (SGD).

 Natural Gradient (NG) \cite{Amari1998a,Amari1997} descent is an optimisation method traditionally motivated from the perspective of information geometry, and works well for many applications \cite{Pascanu2013a,Desjardins2015,Povey2014} as an alternative to SGD. In our previous work \cite{Adnan2017}, it was shown  how the  method when framed in a Hessian Free (HF) styled \cite{Kingsbury2012,Martens2010} optimisation framework is more effective than either variants of SGD or HF for discriminative sequence   training of hybrid HMM-DNN acoustic models. However, the efficacy of this form of NG training fails to extend to DNN models that utilise Rectified Linear Units (ReLUs)  \cite{Vinod2010}.
 
This paper proposes  NGHF, an alternative optimisation method that combines  both  the  NG and HF approaches  to effectively train HMM-DNN models with discriminative sequence criteria.  The NGHF method uses both the direction of steepest descent on a probabilistic manifold and local  curvature information, and is effective for different feed-forward DNN architectures and choices of  activation function. The  method is evaluated on the Multi-Genre Broadcast (MGB) transcription task \cite{Bell2015} and is shown to achieve larger reductions in the Word Error Rate (WER)  for the same number of updates than both NG and HF, as well as lower WERs than SGD.  The  machinery needed to develop this framework relies on the concepts of  manifolds, tangent vectors and directional derivatives from the perspective of information geometry.   An overview of the necessary underlying concepts is provided in \cite{Adnan2018a} but a more in-depth discussion can be found in Amari's  infomation geometry text book \cite{AmariBook}.

The paper is organised as follows. Section \ref{sec:DT}  provides a brief  overview of discriminative sequence training and compares standard derivative based optimisers with NG. Section~\ref{sec:NGHF} formulates the method of NGHF, and  Sec. \ref{sec:GN} discusses the effect of scaling directions with the Gaussian-Newton matrix.  The experimental setup for ASR experiments  is given in Sec. \ref{sec:ES}, with results in Sec.~\ref{sec:E}, followed by the conclusion.
\section{Discriminative Sequence Training \label{sec:DT}}
ASR is a sequence to sequence level classification task where  given an acoustic waveform $\mathcal{O}$, the goal is to produce the correct hypothesis sequence $\mathcal{H}$ through the use of  an inference model $P_{\bm{\theta}}(\mathcal{H}|\mathcal{O})$.
 Let ${X}$  denote  the  parameter manifold.  As different realisations of DNN parameters lead to different probabilistic models  $P_{\bm{\theta}}(\mathcal{H}|\mathcal{O})$,   the manifold  essentially captures the space of all  probability distributions $\mathcal{M}$ that can be generated by a particular model.  The goal of learning is to identify  a viable candidate ${f}(\bm{\theta},\mathcal{O}) \in \mathcal{M}$ that   achieves the greatest reduction in the empirical loss  w.r.t a given risk measure while generalising well to new examples. In ASR,  the WER is the evaluation metric of interest which however corresponds to a discontinuous function of the model parameters.  Hence,  employing such a metric directly  within a empirical risk criterion is not  viable  with  standard derivative  based optimisers.  This forms  the motivation behind the class of Minimum Bayes' Risk (MBR) objective functions \cite{Povey2002,Gibson2006}: 
\vspace{-1mm}
\begin{align}
F_{\rm{MBR}} (\bm{\theta})  &= \frac{1}{R} \sum_r^R \left [ \sum_{\mathcal{H}} P_{\bm{\theta}}(\mathcal{H} |\mathcal{O}^r,\mathcal{M}) L(  \mathcal{H},\mathcal{H}^r)  \right]
\end{align}
where  $(\mathcal{H}^r,\mathcal{O}^r)$ represents the true transcription and feature vectors associated with utterance $r$, and $L$ represents the loss function. In MBR training, instead of minimising the empirical loss for each utterance, the expected loss associated with each utterance in the training set is minimised. Such a function is a smooth function of the DNN model parameters and hence can be optimised by derivative based approaches. In practice, it is not feasible to consider the entire hypothesis space for each utterance without making simplifications to the HMM topology \cite{LMMI}.  The standard approach is to encode confusable hypotheses for with each training utterance using an efficient lattice framework \cite{Woodland2002}.

The premise behind all  derivative based optimisation methods  is Taylor's theorem.  Assuming that  the objective function $F(\bm{\theta})$ is sufficiently smooth, Taylor's 
formula including terms up to the second order
approximates  the local behaviour of the objective  function  by the following quadratic function: 
\vspace{-1mm}
\begin{align}
F(\bm{\theta}_k + \Delta \bm{\theta}) \simeq F(\bm{\theta}_k ) +  \Delta \bm{\theta}^T  \nabla F(\bm{\theta}_k )+ \frac{1}{2} \Delta \bm{\theta}^T H  \Delta \bm{\theta} 
\label{TaylorTheorem}
\end{align}
where $ \Delta \bm{\theta}$ corresponds to any offset within a convex neighbourhood of   $\bm{\theta}_k$ and $H$ is the Hessian. Instead of optimising the objective function directly, second order methods focus on minimising the  approximate quadratic  at each iteration of the optimisation process. The same approach is undertaken by first order methods which only consider the gradient $ \nabla F(\bm{\theta}_k )$, i.e. the first order term. For the class of MBR objective functions,  the gradient associated with the $r$th utterance at time  $t$ w.r.t  the DNN output activations can be shown  to be the component-wise multiplication of the vectors  $\bm{\gamma}^r_t \odot \bm{L}$ \cite{DNNGrad,Matt,Adnan2017}, where $\bm{\gamma}^r_t$  represents the posterior  probability associated with the states (DNN output nodes) at time $t$  and  the entries of $\bm{L}$ correspond to the local phone(state) level  loss associated with these arcs within the consolidated lattice \cite{Gibson2006}.

Solving (\ref{TaylorTheorem})  yields the critical point $\Delta \bm{\theta} = H^{-1}\nabla F(\bm{\theta}_k )$. This corresponds to  a unique minimiser only when the Hessian  $H$ is positive definite.  However, when the choice of models $\mathcal{M}$ is restricted  to DNNs, the Hessian  irrespective of the choice of objective function is no longer guaranteed to be positive definite.  To address this issue,  \cite{Sainath2013c} showed that  by approximating the Hessian with the Gauss-Newton 
\cite{Schraudolph} matrix, solving (\ref{TaylorTheorem})  guarantees an improvement in the training objective function. When the underlying model  corresponds to a discriminative probabilistic model $P_{\bm{\theta}}(\mathcal{H}|\mathcal{O})$, a more natural optimisation method is the method of NG.  In NG,  the updates associated with each iteration correspond to the direction of steepest decent on the probabilistic manifold. In  \cite{Amari1998a,Amari1997,Bottou2016}, such a direction  is shown to equate to the critical point of (\ref{TaylorTheorem}) with the Hessian  replaced by the Fisher Information matrix \cite{AmariBook}.
 

\section{ Formulating  NGHF \label{sec:NGHF}}
 In \cite{Adnan2018a}, it is shown that by deriving Taylor's second order approximation from the perspective of manifold theory,  solving the minimisation problem of  (\ref{TaylorTheorem}) becomes equivalent to solving the following minimisation problem in the tangent space $T_{\bm{\theta}_k} X$:
 \vspace{-2mm}
\begin{align}
 \argmin_{\Delta \bm{\theta} \in T_{\bm{\theta}_k} X} \left[ F(\bm{\theta}_k ) +  \langle \Delta \bm{\theta},  \nabla F(\bm{\theta}_k ) \rangle + \frac{1}{2} \Delta \bm{\theta}^T H  \Delta \bm{\theta}  \right]
\label{IHF}
\end{align}
With such an approach, $  \langle \Delta \bm{\theta},  \nabla F(\bm{\theta}_k ) \rangle$ corresponds to the  inner product between vectors $\nabla F(\bm{\theta}_k)$ and $\Delta \bm{\theta}$ in $T_{\bm{\theta}_k} X$   while  $ \Delta \bm{\theta}^T H  \Delta \bm{\theta}$ represents a linear map $g : \bm{u} \in  T_{\bm{\theta}_k} X \rightarrow \mathbb{R}$.  Since ${X}$ is a manifold,  the inner product endowed on the tangent space  $T_{\bm{\theta}} X$  need not be the identity matrix. The parameter manifold $X$ can be equipped with any form of a  Riemannian metric,  a smooth map that assigns to each $\bm{\theta} \in X$ an inner product  $I_{\bm{\theta}}$  in $T_{\bm{\theta}} X$.  In our previous work \cite{Adnan2017}, it was shown  how for  sequence discriminative training, an ideal choice  of  $I_{\bm{\theta}}$ corresponds to the expectation of the outer product of the Maximum Mutual Information (MMI) \cite{Woodland2002} gradient:
 \vspace{-2mm}
 \begin{align*}
I_{\bm{\theta}}  &=  E_{ P_{\bm{\theta}}(\mathcal{H}|\mathcal{O}) }  \left [ \left( \nabla \mbox { log } P_{\bm{\theta}}(\mathcal{H}|\mathcal{O}) \right ) \left( \nabla \mbox { log } P_{\bm{\theta}}(\mathcal{H}|\mathcal{O}) \right )^T \right ] 
    \end{align*}
 
 As  $I_{\bm{\theta}}$ by definition  is symmetric and positive definite,  it is invertible by the {spectral decomposition theorem}.  If   the  manifold $X$  is now equipped with a Riemannian metric of the form of  $I_{\bm{\theta}}^{-1}$,   then  the dot product $\langle \Delta \bm{\theta},  \nabla F(\bm{\theta}_k ) \rangle$ in (\ref{IHF})  corresponds to $  \Delta \bm{\theta}^T I_{\bm{\theta}}^{-1}  \nabla F(\bm{\theta}_k )$.  Under such a metric, solving the minimising problem by considering only the first two terms in  (\ref{IHF}) equates to performing Natural Gradient on the parameter surface. In  this paper,  the entire quadratic function  of $(\ref{IHF})$ is  considered when solving the minimisation problem in  $T_{\bm{\theta}_k} X$:
 \vspace{-2mm}
\begin{align}
 \argmin_{\Delta \bm{\theta} \in T_{\bm{\theta}_k} X} \left[  F(\bm{\theta}_k ) +  \Delta \bm{\theta}^T I_{\bm{\theta}}^{-1}  \nabla F(\bm{\theta}_k )+ \frac{1}{2} \Delta \bm{\theta}^T H \Delta \bm{\theta} \right] \label{FIHF}
\end{align}
In practice, the expectation of the outer product of the MMI  gradient  is approximated  by its Monte-Carlo estimate $\hat{I}_{\bm{\theta}}$ which is not guaranteed to be positive definite. Thus, its inverse is no longer guaranteed to exist. To address this issue,  \cite{Adnan2018a} provides the derivation of  an  alternative dampened Riemannian metric $ \tilde{I}_{\bm{\theta}}^{-1}$ that  is not only guaranteed to be positive definite but has the feature that its image space is the direct sum of the image  and  the kernel space of the empirical Fisher matrix $\hat{I}_{\bm{\theta}}^{-1}$.

 The  critical point of  (\ref{FIHF}) is $ \Delta \bm{\theta} = H^{-1} {I}_{\bm{\theta}}^{-1}  \nabla F(\bm{\theta}_k )$ and corresponds to an NG direction regularised by multiplication with the inverse of the curvature matrix. In this work, the Hessian is approximated by the Gauss-Newton (GN) Matrix  $G_{\bm{\theta}}$. Section \ref{sec:GN} discusses the particular effect of scaling  directions with $G_{\bm{\theta}}$.
 Computing the individual inverse matrix scalings directly is  expensive  in terms of both computation and storage. Hence in this paper, using the approach highlighted in  \cite{Adnan2017,Kingsbury2012},  the solution of individual inverse matrix scalings is approximated by solving equivalent linear systems using the Conjugate Gradient (CG) \cite{Shewchuk1994b} algorithm. Apart from the obvious computational reasons, the use of CG presents two key advantages: the very first iteration of CG computes an appropriate step size for the direction the algorithm is initialised with. In the case of NGHF, this corresponds to the NG direction. Thus, at each iteration of NGHF, the  resultant update  found  after two runs of CG conforms to  $ \Delta \bm{\theta} = w_1 \Delta \bm{\theta}_{NG} + w_2 \Delta \bm{\theta}_{HF}$, a weighted  combination of  the NG direction and conjugate directions computed using local curvature information. Secondly, when applied to solve the proposed linear system $\tilde{I}_{\bm{\theta}} \Delta \bm{\theta} = \nabla F(\bm{\theta}_k )$, the very first directions explored by the algorithm are guaranteed to be the directions which constitute the image space of $\hat{I}_{\bm{\theta}}$ \cite{NocedalBook}. 
\section{Scaling Directions with the GN Matrix \label{sec:GN}}
When DNN models are employed to solve the inference problem, $G_{\bm{\theta}}$ can be shown to  take the   particular form of  $J_{\bm{\theta}} ^{T} \nabla^2 \hat{L}_{\bm{\theta}} J_{\bm{\theta}}$ where
\begin{itemize}
\item $\nabla^2 \hat{L}_{\bm{\theta}}$ is the Hessian of the loss function w.r.t the DNN  linear output activations with individual entries being functions of $\bm{\theta}$.
\item $J_{\bm{\theta}}$  is the Jacobian of the  DNN output activations w.r.t $\bm{\theta}$.
\end{itemize}
To  keep the notation uncluttered, the dependency on $\bm{\theta}$ will be dropped for the remainder of this section when dealing with the individual factors of  the product $J_{\bm{\theta}} ^{T} \nabla^2 \hat{L}_{\bm{\theta}} J_{\bm{\theta}}$.
 As both $\nabla^2 \hat{L}$ and the product $J^{T} \nabla^2 \hat{L}J$  are real and symmetric, by the {spectral decomposition theorem}: $ J ^{T} \nabla^2 \hat{L}  J \equiv J ^{T} U \Lambda U^T J \equiv V \Sigma V^T$. Under this factorisation, each eigenvector  $\bm{v}_j$  of $J ^{T} \nabla^2 \hat{L} J$ can  be interpreted as a particular weighted sum of the gradients of the output activations of the DNN w.r.t $\theta$. Switching from the standard basis to the basis spanned by $U^T J$,  updates conforming to  directions of steepest descent can be expressed as:
  \vspace{-2mm}
\[ \Delta \theta = \sum_{j}^k \eta  \left( \bm{v}_j^T\nabla F_{\rm{MBR}} \right) U_{j,i} \dfrac{\partial {\tilde{F}}_i}{\partial \theta}\]
where $\eta$ is the learning rate and  $\tilde F $ represents the objective function with the domain constrained to the DNN  output layer. With respect to this basis, scaling with  the inverse of the GN matrix  effectively corresponds to rescaling the steps taken along individual $\bm{v}_j$ by a factor $1/{\mu_j}$:
\vspace{-2mm}
 \begin{align}
  \Delta \theta = \sum_{j}^k \frac{1}{\mu_j}  \left( \bm{v}_j^T\nabla F_{\rm{MBR}} \right) U_{j,i} \dfrac{\partial {\tilde{F}}_i}{\partial \theta}
\end{align}
where $\mu_j$ is  the eigenvalue associated with $\bm{v}_j$ in $V$.   Recall  that   $ \nabla^2 \hat{L} $ can alternatively be presented as $\nabla .( \nabla \tilde F)$. Therefore, eigenvectors $\bm{u}_j$ in $U$ with large eigenvalues correspond to directions that can induce large changes in the gradient of $\nabla \tilde F$.  By establishing a one-to one correspondence between eigenvectors of $ \nabla^2 \hat{L} $  with eigenvectors  of   $J ^{T} A  J$, it can be seen that re-scaling with 
$1/{\mu_j}$ effectively de-weights  back propagation information carried by those paths through the network that can  induce large changes  in  $\nabla \tilde F$. In the context of discriminative training, this ensures that the DNN frame posterior distribution does not  become overly sharp.


\section{Experimental Setup \label{sec:ES}}
The various DNN optimisation approaches were evaluated  on data from the  2015 Multi-Genre Broadcast ASRU challenge task (MGB1) \cite{Bell2015}. In this work, systems were trained using a 200 hour training set\footnote{Note that most results in  \cite{Woodland2015} use a larger 700h training set, stronger language models and other setup differences.}. The  official MGB1 dev.sub set was employed as a validation set and consists of 5.5 hours of audio data. To  estimate the generalisation performance of  candidate models, a separate evaluation test set  dev.sub2 was used.  This comprises roughly of  23 hours of audio from the remaining 35 shows belonging the MGB1 dev.full set. Further details related to the data preparation  can be found in \cite{Woodland2015}.

All experiments were conducted using an extended version of the HTK 3.5 toolkit \cite{Zhang2015,HTKBook15}. This  paper focuses on training  standard fully connected DNNs and Time Delay Neural Networks (TDNNs) \cite{Peddinti} using both ReLU and sigmoid activations. The architecture used for DNNs consisted of five hidden layers each with 1000 nodes.  For TDNNs, the network topology consisted of seven hidden layers each with 1000 hidden units. The context specification used for the various TDNN layers is as follows:  [-2, +2] for layer 1, $\{-1,2\} $ for layer 2, $ \{-3,3\}$ for layer 3, $ \{-7,2\} $ for layer 4 and [0] for the remaining layers. For both models, the output layer consisted of 6k nodes and corresponds to context dependent sub-phone targets formed by conventional decision tree context dependent state tying \cite{statetying}.  For DNNs,  the input to the model  was  produced by splicing together  40  dimensional log-Mel filter bank (FBK) features extended with their delta coefficients across 9 frames to give a 720 dimensional input per frame. While for TDNNs, only the 40 dimensional log-Mel filter bank features were considered. For all experiments, the input features were normalised at the utterance level for mean and at the show-segment level for variance  \cite{Woodland2015}.  

All models were trained using lattice-based MPE training \cite{Povey2002}. Prior to sequence training, the  model parameters were initialised using frame-level CE training. To track the occurrence of  over-fitting due to training criterion mismatch at intermediate stages of sequence training, decoding  was performed  on the validation set  using the same weak pruned biased LM used to create the initial MPE lattices. To evaluate  the generalisation performance of the trained models,  a 158k word vocabulary trigram LM was used to decode  the validation and test set.

\textbf{Training configuration for SGD}:   The  best results with SGD were achieved through  annealing of the learning rates at subsequent epochs. The initial  learning rates  were found through a  grid search. For TDNNs, using momentum was found to yield the best  WERs. 

\textbf{Training configuration for NGHF, NG \& HF}:  The recipe described in \cite{Adnan2017} was used: gradient batches corresponding to roughly 25 hours   and   0.5 hrs of audio were sampled for each CG run.  In all experiments, running each CG run beyond 8 iterations was not found to be advantageous. The  CG computations varied between 18\% to  26\% of the total  computational cost. 

%

\section{Experimental Results \label{sec:E}}
\begin{figure}[t]
\vspace{-2em}
\hspace*{-0.9cm}\includegraphics[scale=0.44]{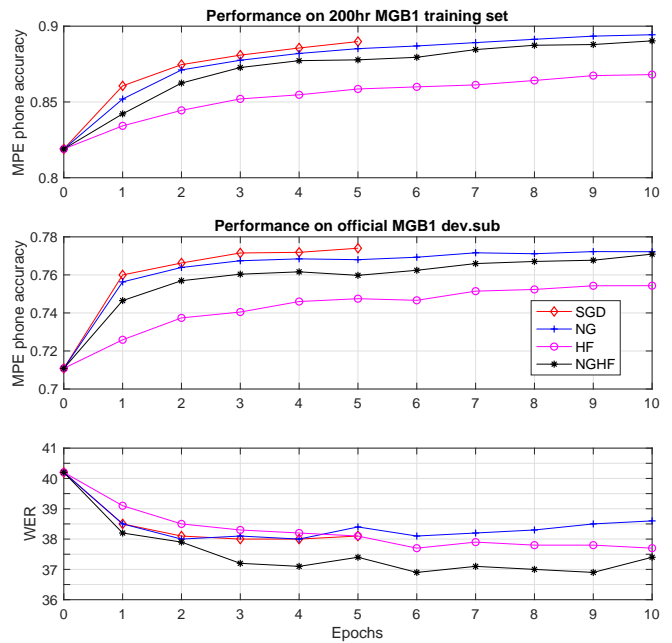}
\vspace*{-3em}
\centering
 \caption{ \label{fig:DNN200} Evolution of  MPE phone accuracy criterion on the training and validation (dev.sub) sets  with ReLU based DNN (top 2 graphs). Also (lower graph) WER with MPE LM on dev.sub as training proceeds.}
\vspace*{-1.5em}
\end{figure}

\vspace*{-0.5em}
Figure \ref{fig:DNN200} compares the performance of various optimisation  methods on training a ReLU based 200hr HMM-DNN model while  Table~\ref{tab:2} shows the performance of these optimisers  for a ReLU based HMM-TDNN system.  
\begin{table}[H]
\tabcolsep=0.1cm
\centering
\begin{tabular}{|c|c|c|cc|c|c|}
\hline
method & \#epochs & \#updates &\multicolumn{2}{c|}{phone acc.} & WER with \\
& & & train  & dev.sub &MPE LM  \\
\hline
CE  & N/A & N/A &  0.870 & 0.754 & 36.9   \\
 \hline\hline
SGD & 4  & 4.64 $\times 10^5$    &0.888   &0.760  &36.4  \\
\hline
 \hline
 NG & 4& 32 &  0.913  & 0.789   &36.2  \\
 \hline
 HF & 4 & 32 & 0.899  & 0.783  & 35.9 \\
 \hline
  NGHF &4  & 32 & 0.911   & 0.791 & \textbf{35.6}  \\
  \hline
 \end{tabular}
\caption{Performance of different optimisers on the  TDNN-ReLU model. WERs on dev.sub.}
\label{tab:2}
\end{table}
\vspace{-1.5em}
It can be seen that among all the optimisers, NGHF  is the most effective in achieving the largest WER reductions on dev.sub with the weak MPE LM. At each iteration, the update produced by the method conforms to $ \Delta \bm{\theta} = w_1 \Delta \bm{\theta}_{NG} + w_2 \Delta \bm{\theta}_{HF}$, a weighted  combination of NG direction and conjugate directions computed using local curvature information.  In Fig.~\ref{fig:DNN200}, it can be seen that by utilising information from both  the KL divergence in the probabilistic manifold and local curvature information, the  proposed method follows a path where optimising the MPE criterion better correlates with achieving reductions in WER. With the ReLU based TDNN as evident from Table~\ref{tab:2}, this same feature can be observed. NGHF achieves better generalisation performance for both the MPE criterion and the WER on the validation set.  To investigate whether these WER reductions hold with stronger LMs  and the relative gains are  not constrained to only ReLU based systems, equivalent systems using sigmoids were trained.  Table~\ref{tab:dev.sub}  compares the performance of  the various optimisers on the validation set with the different models using the 158k LM. 

\begin{table}[H]
\tabcolsep=0.18cm
\begin{tabular}{|c|c|c|c|c|c|}
\hline
Model &  ~~CE~~   & \multicolumn{4}{c|}{MPE}\\
\hline
   & &	SGD   &       NG    & HF   &  NGHF\\
\hline	
DNN-ReLU  &     30.9  &  29.9   &  29.8  & 28.9     &\textbf {28.1} \\

TDNN-ReLU   & 28.6       &  28.5  & 28.7   & 28.1     &\textbf{27.5}\\
\hline
DNN-sigmoid & 31.9     &       29.3   &29.0  & 29.3 &\textbf{29.0}  \\

TDNN-sigmoid   &  28.5    &   27.1    & 26.9 & 27.0  & \textbf{26.6 } \\
\hline
\end{tabular}
\centering
\caption{ \label{tab:dev.sub}WERs on MGB1 dev.sub  with 158k trigram LM.}
\end{table}
\vspace{-1.5em}

 From  Table~\ref{tab:dev.sub}, it can be seen that again models using NGHF achieve the largest reductions in WER. For ReLU based models, NGHF achieves a relative Word Error Rate Reduction (WERR) of 9\%  with the DNN and  4\%  with the TDNN.  Whereas with the sigmoid based models, the method achieves a relative WERR of 6\% with the DNN and 7\%  with  the TDNN.  Compared to SGD,  NGHF is especially effective with the ReLU based models.  For the DNN, the method  achieves a  relative WERR of 6\% over SGD, while with the TDNN the relative WERR is 4\%.

Finally, the generalisation performance of the trained models were estimated by performing Viterbi decoding on dev.sub2 using 158k LM. Results are shown in Table~\ref{tab:gen}. 

\begin{table}[H]
\begin{tabular}{|c|c|c|c|c|c|}
\hline
Model &  CE   & \multicolumn{4}{c|}{MPE }\\
\hline
   & &	SGD   &       NG    & HF &   NGHF\\
\hline
DNN-ReLU  &    32.3           &  31.9         &   31.4 & 30.6     & \textbf{29.8} \\
TDNN-ReLU  &  30.6    & 29.8           & 30.6 & 29.6 & \textbf{29.3} \\
\hline
DNN-sigmoid   & 33.2     &       30.8     & 30.5 & 30.9 &\textbf{30.5}  \\
TDNN-sigmoid  &  29.9    & 28.6 &28.2&28.4  & \textbf{27.9} \\
\hline
\end{tabular}
\centering
\caption{ WERs  on MGB1 dev.sub2 with 158k trigram LM. }
\label{tab:gen}
\end{table}
\vspace{-1.5em}

It can be observed that as before the model trained with NGHF achieves the largest reductions in WER. With sigmoid   based models, the  method can be seen to be achieve WERR reductions of 8\% with the DNN and 7\% with the TDNN. Over standard SGD, the proposed method achieves relative WERR  of  7\% with  the ReLU-DNN model and a 2\%   with the ReLU based TDNN model.  


\subsection{Investigating overfitting due  to Criterion Mismatch}
\begin{figure}[t]
\hspace*{-0.6cm}\includegraphics[scale=0.37]{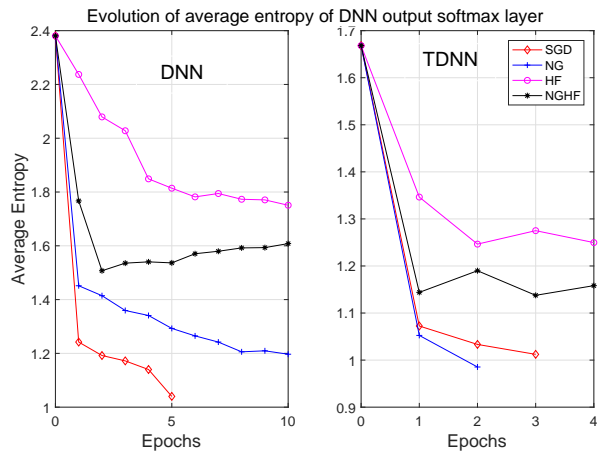}
\vspace*{-1.5em}
\centering
\caption{ Evolution of  average entropy of DNN output activations during MPE training with ReLU based DNN models. Left graph is for DNNs and the right graph for TDNNs.}
\label{fig:entropy}
\vspace*{-1.5em}
\end{figure}

ReLU based systems failed   to achieve  similar WERRs as sigmoid  based systems  from sequence training   with either SGD or NG.
After a few epochs, improvements made on the MPE criterion failed to correlate with lower WERs. This effect was particularly noticeable with the TDNN model. It was observed that this emergence of criterion mismatch is  correlated with the sharp decrease in the average entropy of  DNN output frame posteriors (Fig.~\ref{fig:entropy}). MBR training broadly speaking tries to concentrate probability mass: a sufficiently flexible model trained to convergence with MBR will assign a high probability to those hypotheses that have the  smallest loss. This means that during the course of training,  the posterior distribution  of states $\bm{\gamma}^r_t(i)$ associated with high local losses $\bm{L}(i)$ is gradually reduced.  With hyper-parameters such as LM and acoustic model scale factors fixed, this sharp decrease in the DNN output entropy directly reflects that this distribution $\bm{\gamma}^r_t$ is becoming overly sharp,
which was found to be detrimental to the decoding performance.

  With sigmoid models, the criterion mismatch was found to be  less severe when using first order methods (Table~\ref{tab:dev.sub}). It was observed that the average entropy of  the DNN frame posteriors with sigmoid models was much larger at the start of sequence training. This is expected as   ReLUs allow a better  flow of gradients during back-propagation resulting in  better  CE trained discriminative models. However,  as observed in Fig.~\ref{fig:entropy}, this also results in sharper frame posterior distributions. From Fig.~\ref{fig:entropy}, it can also be seen  how   scaling with the GN matrix  helps  regularise the  entropy of the DNN frame posteriors.  When compared to HF,  the proposed NGHF approach is better in finding a balance between improving  the MPE criterion and regularising the entropy changes of the DNN frame posteriors.


To improve  generalisation performance,  techniques such as  dropout \cite{Srivastava} and L2 regularisation with SGD sequence training were investigated. However, both of these techniques  were unsuccessful in alleviating  this overfitting due to criterion mismatch. To improve NG training,  the use of Tikhonov damping as advised by Martens \cite{Martens2010} was also investigated to help regularise  the NG updates. Taking comparatively more conservative steps along conjugate directions at the expense of slower learning was observed to regulate the decrease in the average entropy of DNN frame posteriors. However,  the  damped optimiser failed to achieve  better convergence.

\section{Conclusion}

This paper has introduced a new  optimisation method  to effectively train HMM-DNN models with discriminative sequence training. The efficacy of the method has been shown to be agnostic with respect to both the choice of feed forward architecture and choice of DNN activation functions. When applied within a HF styled optimisation framework, the proposed methods enjoys the same benefits as HF but leads to better convergence than NG, HF and SGD.  Future work  will involve extending the proposed framework to  training DNN architectures with recurrent topologies.
\bibliographystyle{IEEEtran}

\begin{thebibliography}{10}

\bibitem{Hinton2012}
G.~Hinton, L.~Deng, D.~Yu, G.~Dahl, A.~Mohamed, N.~Jaitly, A.~Senior,
  V.~Vanhoucke, P.~Nguyen, T.N.~Sainath \& B~Kingsbury,
\newblock ``{Deep Neural Networks for Acoustic Modeling in Speech Recognition:
  The Shared Views of Four Research Groups}",
\newblock {\em IEEE Signal Processing Magazine}, vol. 29, no. 6, pp.
  82--97, 2012.
  
 \bibitem{Amari1998a}
S.~Amari,
\newblock ``{Natural Gradient Works Efficiently in Learning}",
\newblock {\em Proc. Neural Information Processing Systems (NIPS)}, 1998.

\bibitem{Amari1997}
S.~Amari,
\newblock ``{Neural Learning in Structured Parameter Spaces--Natural Riemannian Gradient}",
\newblock  {\em Proc. Advances in Neural Information Processing
  Systems (NIPS)}, 1997.
  
  
  \bibitem{Pascanu2013a}
R.~Pascanu \&Y.~Bengio,
\newblock ``{Revisiting Natural Gradient for Deep Networks}",
\newblock {\em arXiv preprint arXiv:1301.3584}, Jan. 2013.

\bibitem{Desjardins2015}
G.~Desjardins, K.~Simonyan \& R.~Pascanu,
\newblock ``{Natural Neural Networks}",
\newblock {\em Proc. Advances in Neural Information Processing Systems
  (NIPS)}, 2015.

  \bibitem{Povey2014}
D.~Povey, X.~Zhang \& S.~Khudanpur,
\newblock ``{Parallel Training of DNNs with Natural Gradient and Parameter
  Averaging}",
\newblock {\em arXiv preprint arXiv:1410.7455}, Oct. 2014.


  \bibitem{Adnan2017}
  A.~Haider \& P.C.~Woodland,
 \newblock ``{Sequence Training of DNN Acoustic Models with Natural Gradient}",
\newblock {\em IEEE Workshop on Automatic Speech Recognition and
 Understanding (ASRU)}, 
Dec. 2017.
    
\bibitem{Kingsbury2012}
B.~Kingsbury, T.N. Sainath \& H.~Soltau,
\newblock ``{Scalable Minimum Bayes Risk Training of Deep Neural Network
  Acoustic Models using Distributed Hessian-Free Optimisation}",
\newblock  {\em Proc. Interspeech}, 2012.

\bibitem{Martens2010}
J.~Martens,
\newblock ``{Deep Learning via Hessian-Free Optimisation}",
\newblock {\em Proc. International Conference on Machine Learning (ICML)},
  2010.

\bibitem{Vinod2010}
V.~Nair \& G.~Hinton,
\newblock ``{Rectified Linear Units Improve Restricted Boltzmann Machines}",
\newblock {\em Proc. International Conference on Machine Learning (ICML)},
2010.


\bibitem{Bell2015}
P.~Bell, M.~Gales, T.~Hain, J~Kilgour, P.~Lanchantin, X~Liu, A~McParland,
  S.~Renals, S.~Saz, N~Wester \& P.~ Woodland,
\newblock ``{The MGB Challenge: Evaluating Multi-Genre Broadcast Media
  Recognition}'',
\newblock  {\em Proc. IEEE Workshop on Automatic Speech Recognition and Understanding (ASRU)}, 2015.


\bibitem{Adnan2018a}
A.Haider,
\newblock ``{A Common Framework for Natural Gradient  and Taylor based Optimisation using Manifold Theory}",
\newblock {\em arXiv preprint}
 2018.

 \bibitem{AmariBook}
S.~Amari,
\newblock ``{ Information Geometry and its Applications}",
\newblock{\em Springer 2016}, 
ch. 1, pp. 3-27.

\bibitem{Povey2002}
D.~Povey \& P.C. Woodland,
\newblock ``{Minimum Phone Error and I-smoothing for Improved Discriminative
  Training},''
\newblock  {\em Proc. ICASSP}, 2002.


\bibitem{Gibson2006}
M.~Gibson  \& T.~Hain,
\newblock ``{Hypothesis Spaces for Minimum Bayes' Risk Training in Large
  Vocabulary Speech Recognition}",
\newblock {\em Proc. Interspeech}, 2006.

\bibitem{LMMI}
D.~Povey, V.~ Peddinti, D.~Galvez,, P.~Ghahremani, V.~Manohar, X.~Na,Y.~Wang \& S.Khudanpur,
\newblock ``{Purely Sequence-Trained Neural Networks for ASR based on Lattice-Free MMI}'',
\newblock { \em Proc. Interspeech}, 2016.
 
 \bibitem{Woodland2002}
P.C.~ Woodland \& D.~Povey,
\newblock ``{Large Scale Discriminative Training of Hidden Markov Models for Speech Recognition}",
\newblock {\em Computer Speech and Language}, vol. 16. pp. 25-47, 2002.

 
 \bibitem{DNNGrad}
 K.~Vesely, A.~Ghosal, L.~Burget \& D.~Povey,
 \newblock ``{Sequence-Discriminative Training of Deep Neural Networks}",
 \newblock{\em Proc. Interspeech},  2013.
 
 \bibitem{Matt}
 M.~Shannon,
 \newblock ``{Optimizing Expected Word Error Rate via Sampling for Speech Recognition"},
 \newblock{\em Proc. Interspeech}, 2017. 
 
 
 
 \bibitem{Sainath2013c}
T.N. Sainath, B.~Kingsbury \& H.~Soltau,
\newblock ``{Optimization Techniques to Improve Training Speed of Deep Neural
  Networks for Large Speech Tasks}",
\newblock {\em  IEEE Trans. on Audio, Speech, and Language Processing}, vol. 21, no. 11, pp. 2267--2276, 2013.


 
\bibitem{Schraudolph}
N.~Schraudolph,
\newblock``{Fast Curvature Matrix-Vector Products for Second-Order Gradient Descent}",
\newblock{\em Proc.  Neural Information Processing Systems (NIPS)},
2002.

\bibitem{Bottou2016}
L.~Bottou, F.~Curtis \& J.~Nocedal,
\newblock {\em {Optimization Methods for Large-Scale Machine Learning}},
\newblock arXiv preprint arXiv:1606.04838, 2016.



\bibitem{Shewchuk1994b}
J.R. Shewchuk,
\newblock ``{An Introduction to the Conjugate Gradient Method Without the
  Agonizing Pain}", 1994.
  
 \bibitem{NocedalBook}
  J.~Nocedal \& S.~Wright,
 \newblock ``{Numerical Optimization}",
 \newblock {\em  2nd Edition, Springer series in Operations Research},
 ch. 5, pp. 115.
  

\bibitem{Woodland2015}
P.C.~Woodland, X.~Lui, Y.~Qian, C.~Zhang, M.J.F.~Gales, P.~Karanasou,
  P.~Lanchantin \& L.~Wang,
\newblock ``{Cambridge University Transcription Systems for the Multi-Genre
  Broadcast Challenge}",
\newblock {\em Proc. IEEE Workshop on Automatic Speech Recognition and Understanding (ASRU)}, 2015.

\bibitem{Zhang2015}
C.~Zhang \& P.C.~Woodland,
\newblock ``{A General Artificial Neural Network Extension for HTK}",
\newblock {\em Proc. Interspeech}, 2015.

\bibitem{HTKBook15}
S.J.~Young, G.~Evermann, M.J.F.~Gales, T.~Hain, D.~Kershaw, X.~Liu, G.~Moore,
  J.J.~Odell, J.~Ollason, D.~Povey, A.~Ragni, V.~Valtchev, P.C.~Woodland \&
  C.~Zhang,
\newblock {\em The HTK Book (for HTK 3.5)},
\newblock Cambridge University Engineering Department, 2015.

\bibitem{Peddinti}
V.~Peddinti, D.~Povey and S.Khudanpur
\newblock ``{A Time Delay Neural Network Architecture for Efficient Modeling of Long Temporal Contexts}"
\newblock{\em Proc. InterSpeech}
 2015.
  
 \bibitem{statetying}
 S.J.~Young, J.J.~Odell \& P.C.~Woodland
 \newblock ``{Tree-Based State Tying for High Accuracy Acoustic Modelling}",
 \newblock {\em Proc  HLT},
 1994.
   

\bibitem{Srivastava}
N.~Srivastava, G.~ Hinton, A.~Krizhevsky, I.~Sutskever \& R.Salakhutdinov,
 \newblock ``{Dropout: A Simple Way to Prevent Neural Networks from Overfitting}",
 \newblock{ \em The Journal of Machine Learning Research},
 vol. 15, no. 1,  pp. 1929-1958, 2014.
  

\end{thebibliography}


\end{document}